\RequirePackage[hyphens]{url}
\documentclass[letterpaper, 10 pt, conference]{ieeeconf}
\IEEEoverridecommandlockouts %
\makeatletter
\let\MYcaption\@makecaption
\makeatother
\usepackage[font=footnotesize]{subcaption}
\makeatletter
\let\@makecaption\MYcaption
\makeatother

\usepackage{epsfig}
\usepackage{endnotes}

\usepackage[utf8]{inputenc}
\usepackage[utf8]{luainputenc}

\usepackage{ifthen}

\usepackage[table]{xcolor}
\usepackage{diagbox}

\usepackage{nicefrac}

\usepackage{hhline}

\usepackage{tabularx}

\newcounter{myctr}
\newenvironment{mylist}{\begin{list}{(\textbf{\arabic{myctr}})}
{\usecounter{myctr}
\setlength{\topsep}{0mm}\setlength{\itemsep}{0mm}
\setlength{\parsep}{0mm}
\setlength{\itemindent}{0mm}\setlength{\partopsep}{0mm}
\setlength{\labelwidth}{-2mm}
\setlength{\leftmargin}{0mm}}}{\end{list}}

\newif\ifcomments
\commentsfalse
\ifcomments
\usepackage[textsize=small,color=lightgray]{todonotes}
\else
\usepackage[disable]{todonotes}
\fi

\usepackage{textcomp}

\usepackage{pgfplots}
\pgfplotsset{compat=1.13}

\let\labelindent\relax
\newcommand{\subparagraph}{}
\usepackage{enumitem}

\newcommand*\bubble[1]{\tikz[baseline=(char.base)]{
            \node[shape=circle,scale=0.8,draw,inner sep=2pt,fill=black,
            text=white] (char)
            {#1};}}

\usepackage{etoolbox}
\makeatletter
\patchcmd{\ttlh@hang}{\parindent\z@}{\parindent\z@\leavevmode}{}{}
\patchcmd{\ttlh@hang}{\noindent}{}{}{}
\makeatother

\usepackage{microtype}

\usepackage{balance}

\usepackage{datetime}

\usepackage{listings}

\usepackage{tikz}
\usetikzlibrary{calc}
\usetikzlibrary{trees}
\usetikzlibrary{positioning}
\usetikzlibrary{arrows}
\usetikzlibrary{arrows.spaced}
\usetikzlibrary{arrows.meta}
\usetikzlibrary{chains, matrix}
\usetikzlibrary{shapes, shapes.geometric, shapes.symbols, shapes.multipart}
\usetikzlibrary{decorations.markings, decorations.pathreplacing, decorations.pathmorphing, decorations.text, decorations.shapes}
\usetikzlibrary{backgrounds}
\usetikzlibrary{patterns}
\usetikzlibrary{fit}

\usepackage{amsfonts}
\usepackage{amsmath}
\usepackage{amssymb}
\usepackage{pifont}

\usepackage{mathptmx}

\usepackage{xfrac}

\usepackage[hyphens]{url}

\usepackage{cite}

\usepackage{rotating}
\usepackage{booktabs}
\usepackage{xparse}

\usepackage{outlines}

\usepackage{xspace}
\usepackage{framed}

\usepackage{setspace}

\usepackage{adjustbox}

\newboolean{draft}
\setboolean{draft}{false}

\ifcomments
\paperwidth=\dimexpr \paperwidth + 5cm\relax
\oddsidemargin=\dimexpr\oddsidemargin + 2.5cm\relax
\evensidemargin=\dimexpr\evensidemargin + 2.5cm\relax
\marginparwidth=\dimexpr \marginparwidth + 2.5cm\relax
\fi

\newcommand{\myparagraph}[1]{\noindent{\bfseries #1}}

\NewDocumentCommand{\rot}{O{45} O{1em} m}{\makebox[#2][l]{\rotatebox{#1}{#3}}}

\usepackage{comment}

\usepackage{graphicx,color}

\DeclareMathAlphabet{\mathcal}{OMS}{cmsy}{m}{n}

\newcommand{\one}{({\em i}\/)}
\newcommand{\two}{({\em ii}\/)}
\newcommand{\three}{({\em iii}\/)}

\newcommand{\av}{AV}

\usepackage{multirow}

{\end{list}}

{\end{list}}

\makeatletter
\let\NAT@parse\undefined
\makeatother
\usepackage{hyperref}
\hypersetup{
    colorlinks,
    linkcolor={red!50!black},
    citecolor={blue!50!black},
    urlcolor={blue!80!black}
}

\usepackage[capitalize]{cleveref}
\crefformat{section}{#2\S#1#3}
\crefformat{subsection}{#2\S#1#3}
\crefname{section}{\S}{\S}

\title{Leveraging Cloud Computing to Make Autonomous Vehicles Safer}

\author{
    Peter Schafhalter$^{1}$
    \and
    Sukrit Kalra$^{1}$
    \and
    Le Xu$^{2}$
    \and
    Joseph E. Gonzalez$^{1}$
    \and
    Ion Stoica$^{1}$
    \thanks{$^{1}$ UC Berkeley   $^{2}$ UT Austin}
    }
\begin{document}

\date{}
\maketitle

\begin{abstract}
    The safety of autonomous vehicles (\av{}s) 
    depends on their ability to perform complex 
    computations on high-volume sensor data in a timely manner.
    Their ability to run these computations with state-of-the-art models is 
     limited by the processing power and slow update cycles of their onboard hardware.
    In contrast, cloud computing offers the ability to burst computation to vast amounts of the latest generation of hardware. 
    However, accessing these cloud resources requires traversing wireless networks that are often considered to be too unreliable for real-time \av{} driving applications. 

    Our work seeks to harness this unreliable cloud to enhance the accuracy of
    an \av{}'s decisions, while ensuring that it can always fall back to its
    on-board computational capabilities.
    We identify three mechanisms that can be used by \av{}s to safely leverage
    the cloud for accuracy enhancements, and elaborate why current execution
    systems fail to enable these mechanisms.
    To address these limitations, we provide a system design based on the
    speculative execution of an \av{}'s pipeline in the cloud, and show the
    efficacy of this approach in simulations of complex real-world scenarios
    that apply these mechanisms.
\end{abstract}

\section{Introduction} \label{s:introduction}
Autonomous Vehicles (\av{}s) are predicted to make our roads significantly 
safer by eliminating the vast majority of traffic accidents that contributed 
to $38,824$ fatalities in $2020$ in the U.S. 
alone~\cite{national-high-safety-2020}.
However, to fulfill their potential, \av{}s must surpass the safety levels of
human drivers
and handle a diverse set of challenging driving 
scenarios~\cite{erdos,wang2020safety}.
Prior works~\cite{gog2021pylot,wang2022adept,dreossi2019verifai} have 
demonstrated the inextricability of driving safety with the accuracy of the 
\av{}'s computational pipeline, which is governed by the following three 
characteristics:

\myparagraph{High-fidelity data.}
Integrating data from a wide array of high-fidelity sensors can
significantly enhance computational accuracy~\cite{tesla-hardware-4}.
As a result, newer generations of \av{}s seek to expand the number and quality
of onboard sensors, such as cameras, LiDARs, and radars.
For example, Waymo's $5$\textsuperscript{th}-generation \av{}s feature
$29$ cameras as opposed to $19$ in the $4$\textsuperscript{th}-generation.
These additional sensors form a new peripheral vision system that
bolsters safety by providing the \av{} with higher quality data at higher
frequencies~\cite{waymo-4th-gen-specs,waymo-av-sensors}.

\myparagraph{State-of-the-art computation.} The increasing amounts of
available data need to be processed by state-of-the-art algorithms and models
to ensure highly-accurate results.
However, more accurate algorithms and models often come at the cost of
increased parameter sizes and more floating point operations 
(FLOPs)~\cite{liu2021swin,li2022exploring,liu2022convnet}.
Notably, the increased scale of modern deep neural networks (DNNs) forms the
``primary ingredient`` in achieving state-of-the-art 
accuracy~\cite{zhai2022scaling}.
Innovative solutions to reduce the computational requirements of 
DNNs~\cite{tan2021efficientnetv2,tan20efficientdet}
have usually resulted in reduced accuracy,
which hampers the \av{}'s safety in complex environments.

\myparagraph{Timely results.} 
To surpass the safety levels of human drivers, an \av{} must respond more 
quickly than humans~\cite{driving-michigan}, whose reaction times
vary from $390$ ms~\cite{wolfe2020rapid} to 
$1.2$ s~\cite{johansson1971drivers} depending on factors such as road
conditions, driver attentiveness, age, etc.
To achieve these reaction times and ensure safety in challenging driving
scenarios, the \av{}'s computational pipeline must operate in ``real-time'' to 
meet latency goals when executing state-of-the-art computation on 
high-fidelity data.

Thus, to drive safely, \av{}s must produce
highly-accurate and timely results using state-of-the-art algorithms and models 
that consume high-fidelity sensor data.
The combination of these characteristics requires \av{}s to exploit the compute
capabilities of cutting-edge hardware.
However, the deployment of such hardware in an \av{} is constrained by its
cooling, power, and stability requirements~\cite{driving-michigan}.
For example, the DRIVE platform~\cite{nvidia-drive}, NVIDIA's flagship 
hardware for \av{}s, is updated every 3 years.
Its most recent iteration, the DRIVE Orin~\cite{nvidia-orin}, was put into
production vehicles in 2023 and its successor, the DRIVE
Atlan~\cite{nvidia-drive-atlan}, is slated for release in
2026~\cite{nvidia-orin-hyperion}.
Moreover, upgrading the hardware of previously-deployed \av{}s is often
infeasible due to the cost and complexity involved with a 
recall~\cite{tesla-hardware-retrofit}.
As a result, modern \av{}s are forced to trade-off accuracy, and hence, 
safety, for computational resources and timely results, by either reducing the 
amount of data fused from multiple sensors, or deploying algorithms and models 
that require a lower number of parameters and FLOPs~\cite{erdos}.

In light of this fundamental mismatch between the pace of development of 
compute technologies with the update cycles of 
vehicles~\cite{nytimes-pace-mismatch}, we propose to 
augment the computational resources in an \av{} with the
compute capabilities of the cloud.
Cloud computing platforms provide the illusion of infinite computing 
resources~\cite{fox2009above}, and enable low-cost access to state-of-the-art
hardware~\cite{nvidia-gpu,google-tpu,aws-inferentia}.
In contrast to the $3$-year update cycle in \av{}s, the
hardware and software in the cloud is frequently 
updated~\cite{aws-instances,john-wilkes-talk}.
For \av{}s, the cloud enables the deployment of compute-intensive,
rapidly-evolving algorithms and models which can exploit state-of-the-art
hardware without requiring complex recalls for hardware and software
updates~\cite{tesla-recall,toyota-recall-14}.
As a result, \av{}s can enhance safety by executing highly-accurate
algorithms and models on powerful cloud resources to provide timely results.

Despite the potential benefits of the cloud, its practical use in
\av{}s has remained largely unexplored due to concerns regarding 
the limited bandwidth of the network available on \av{}s and the high,
often unpredictable latency of cloud systems.
In this paper, we evaluate the efficacy of augmenting the computational 
resources on-board \av{}s with the unreliable resource pool of the cloud.
Specifically, we seek to enhance the accuracy of an \av{}'s decisions by 
harnessing the capabilities of the cloud, while ensuring that the \av{} can
always fall back to its on-board computational capabilities in order to
strictly exceed its current safety standards.
To this effect, the paper makes the following key contributions:
\begin{mylist}
    \item{} We identify three mechanisms that can be used by \av{}s to 
    leverage the cloud for safety enhancements (\cref{s:techniques}).
    \item{} We identify limitations in the ability of current execution 
    systems to achieve these safety-enhancement mechanisms 
    (\cref{s:related-work}), and propose a system design based on speculative 
    execution of computation in the cloud that enables their efficient 
    execution and maximizes accuracy (\cref{s:design}).
    \item{} We evaluate the efficacy of our design under current cloud 
    and network capabilities using the three mechanisms on complex scenarios 
    from the NHTSA crash scenarios (\cref{s:evaluation}).
\end{mylist}

We argue that the network and compute trends are in favor of the
feasibility of our approach, and discuss specific challenges that must be
resolved by the community to enable \av{}s to reap the complete safety
benefits of the cloud (\cref{s:discussion}).

\begin{figure}[t!]
  \medskip
    \begin{center}
        \includegraphics[width=0.9\columnwidth]{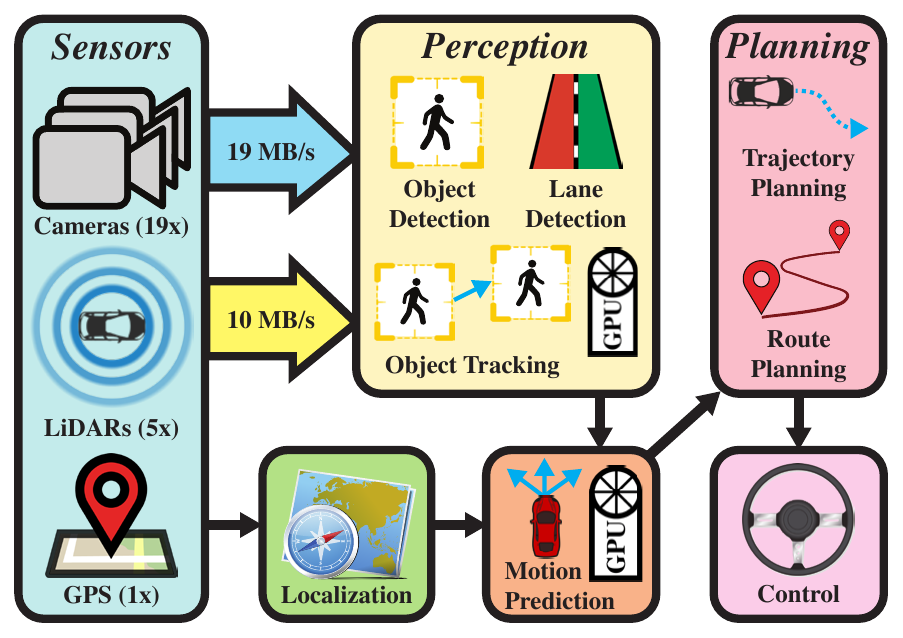}
    \end{center}
    \caption{\textbf{A modular \av{} pipeline} with multiple sensors feeding
    data to the perception, prediction, planning and control modules.}
    \label{fig:self-driving-diagram}
    \vspace{-1.5em}
\end{figure}

\section{Background \& Motivation}
\label{s:background}

The prevalent design of an \av{}'s computation is
a modular multi-stage pipeline (\cref{fig:self-driving-diagram}) 
\cite{cruise-report,ntsb-uber,gog2021pylot,driving-michigan,
autoware,apollo-baidu-new} where the 
inputs from the vast array of sensors (e.g., cameras, LiDARs, radars, etc.) are
processed by the perception and localization modules.
The perception module detects and tracks nearby objects of interest, such as
pedestrians, vehicles, traffic lights, etc.
By fusing the perception results with the location from the localization
module, the \av{} pipeline constructs an internal representation of the
environment.
The prediction module forecasts the behavior of non-static objects
(i.e., pedestrians, vehicles) which enables the planning
module to compute a safe and comfortable trajectory for the \av{} to follow.
Finally, the control module converts the trajectory plan into steering and
acceleration commands which are applied to the \av{}.

\begin{table}[!t]
  \medskip

\captionsetup{size=footnotesize}
    \caption{runtime disparity between av hardware and cloud hardware} \label{t:model-detection}
\setlength\tabcolsep{0pt} %
\scriptsize\centering

\begin{tabular*}{\columnwidth}{@{\extracolsep{\fill}}lccc}
\toprule
    \centering \textbf{Model}  & \multicolumn{2}{c}{\textbf{Runtime [ms]}} & \textbf{Speedup}\\
      & \textbf{Orin} & \textbf{A100} & \\
\midrule
    DETR-ResNet-50 & $301.7$   & $102.2$    & $2.95{\times}$     \\
    DETR-ResNet-101 &  $407.7$  & $118.2$     & $3.45{\times}$     \\
    DETR-ResNet-101-DC & $859.2$   & $146.6$     & $5.86{\times}$     \\
    DINO-SWIN-Tiny & $722.1$ & $90.1$ & $8.01{\times}$ \\
    DINO-SWIN-Small & $903.5$ & $107.1$ & $8.43{\times}$ \\
    DINO-SWIN-Large & $1529.9$ & $180.8$ & $8.46{\times}$ \\
\bottomrule
\end{tabular*}

\vspace{-1.5em}
\end{table}

To compute safe control commands, \av{}s depend on
high-fidelity data from the sensors and timely results from its algorithms and models
(see \cref{s:introduction}).
However, due to limitations in the power of the hardware relative to the amount of
sensor data generated, \av{} developers must make decisions about
about \emph{which} data to process using \emph{what} algorithms and \emph{when} 
to return results to ensure the safety of the vehicle.
For example, Baidu's Apollo \av{}~\cite{apollo-baidu-new} only utilizes
specialized cameras for detecting traffic lights when the \av{} is notified
of the presence of a nearby traffic light via a pre-computed map of the
city~\cite{apollo-tl}.
While this strategy makes efficient use of the hardware resources on-board, the 
strategy fails to detect temporary traffic signals installed by construction or
emergency vehicles, which may lead to unsafe driving scenarios.
Augmenting \av{} compute resources with the cloud will enable more
comprehensive data processing which avoids discarding data due to
resource limitations, thus improving safety without impacting
the critical-path computation.

After deciding \emph{which} data to process, \av{}s must decide
\emph{what} algorithms and models to execute. 
As discussed in \cref{s:introduction}, \av{}s must often compromise
the accuracy of its algorithms and models due to resource constraints
stemming from the hardware available on-board.
Even after accounting for the slow upgrade cycle which often puts the hardware
in \av{}s several revisions behind the state-of-the-art in the cloud,
the hardware available in the cloud is much more powerful than the hardware
available in \av{}s, even if the architecture and the revision are the same.
For example, the NVIDIA DRIVE Orin platform~\cite{nvidia-orin}, which is slated
to be available in production vehicles in 2023 and is the latest revision
supported by Baidu's Apollo \av{}~\cite{apollo-baidu-hw}, uses NVIDIA's Ampere
microarchitecture to deliver a performance
of $5.2$ FP$32$ TFLOPs~\cite{nvidia-drive-orin-spec}.
The equivalent Ampere cloud GPU is the NVIDIA A$100$, which was released in 2020
and delivers a performance of $19.5$ FP$32$ TFLOPs~\cite{nvidia-a100-spec}, a 
$3.75{\times}$ increase over the DRIVE Orin.
\cref{t:model-detection} measures the effects of this disparity by executing 
open-source implementations of
DETR~\cite{carion2020end,detr-huggingface} and
SWIN~\cite{liu2021swin,zhang2022dino}, two state-of-the-art
vision transformers which have been adapted for object detection,
atop both the NVIDIA Orin and the A$100$. 
We observe a significant speedup of up to $8.4{\times}$ by executing the same
model on the same input on a cloud GPU.
Similar trends have been shown by prior work for other safety-critical algorithms
applicable to \av{}s, such as motion planning~\cite{fogros}.

Finally, an \av{} must decide by \emph{when} to compute the control commands
at the end of the pipeline, thus mandating a deadline on the computation being
executed.
For example, to surpass human driving in safety, the pipeline must execute faster
than human reaction times.
In addition, prior
works~\cite{streaming-perception,context-aware-streaming,erdos} indicate
this deadline varies widely according to the environment around the \av{}.
Intuitively, driving slowly on a crowded urban street may tolerate a 
lax deadline for the computation to finish, whereas swerving to prevent an accident 
on the freeway requires the \av{} to respond 
quickly~\cite{claussmann2019review,alcon2020timing}.
This variability in the required response time of the computation further
complicates the deployment of more accurate, but slower algorithms and models
since they cannot meet tighter deadlines~\cite{erdos}.

\section{Enhancing Safety Using The Cloud}
\label{s:techniques}

\begin{figure}[t!]
    \medskip
    \begin{center}
        \includegraphics[width=0.9\columnwidth]{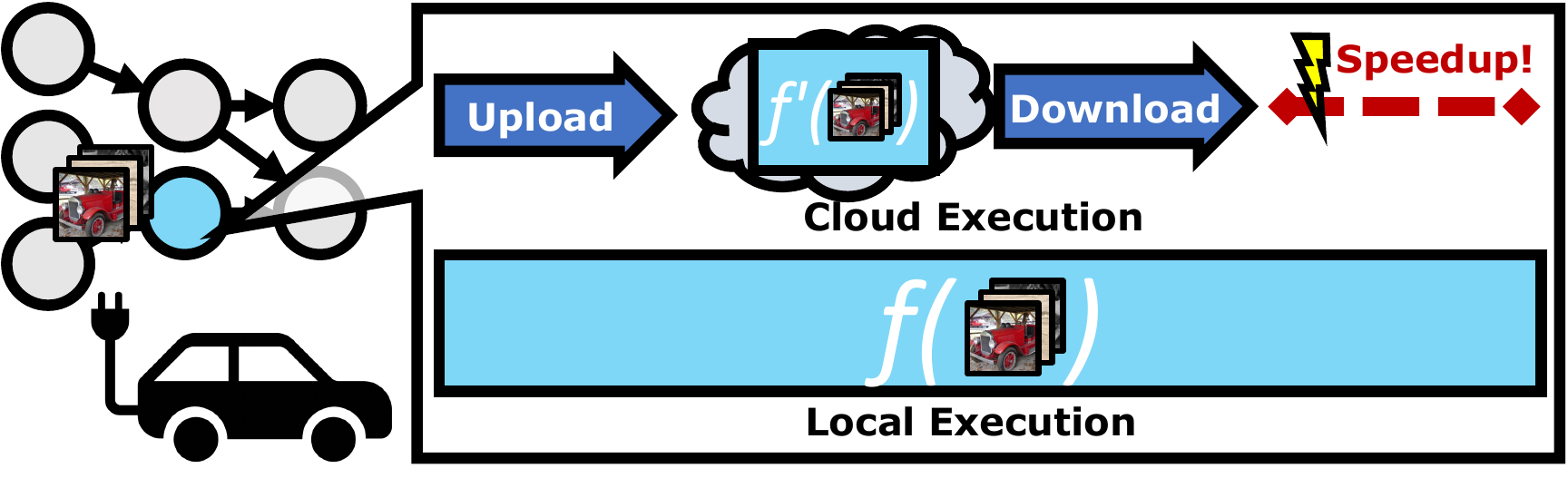}
    \end{center}
    \caption{\textbf{An arbitrage opportunity} afforded by exploiting 
    state-of-the-art hardware to execute higher-accuracy models at a lower 
    latency.}
    \label{fig:cloud-arbitrage}
    \vspace{-1.5em}
\end{figure}

Conventional wisdom suggests that the latency and availability of cellular network connections 
makes using the cloud on the critical-path of the computation infeasible~\cite{kehoe2015survey}.
However, \cref{t:model-detection} presents an arbitrage opportunity whereby
an \av{} could potentially return more accurate results faster by exploiting the computational power of the hardware in the cloud (see \cref{fig:cloud-arbitrage}).
We argue that \av{}s should instead take a best-effort speculative approach to
leverage the cloud when it is available and utilize reliable fallback mechanisms
that use on-board computation when the cloud is not immediately accessible.
We discuss three such mechanisms that enable a best-effort augmentation of an
\av{}'s safety below:

\myparagraph{Higher-accuracy models.} \av{}s can selectively offload data
    from their input sensors to the cloud, allowing higher accuracy models to
    be executed in the cloud.
    For example, in \cref{t:model-detection}, an
    \av{} can choose to execute DETR-ResNet-$101$-DC in the cloud on its
    camera data, which provides a ${\sim}3$-point increase in average precision
    over DETR-ResNet-$50$~\cite{carion2020end} and executes faster.
    Thus, while an \av{} with an end-to-end deadline of $500$ ms can only 
    execute DETR-ResNet-$50$ on its local hardware, it can optimistically 
    exploit the accuracy of all models in \cref{t:model-detection} 
    when the latency to the cloud is low.

\myparagraph{Accurate environment representation.}
    While the earlier mechanism significantly enhances an \av{}'s ability to process
    sensor data and understand its surroundings, this mechanism does not apply
    to obstacles obscured by the \av{}'s blind spots.
    To improve safety in these scenarios, AVs can share their locations computed
    by the localization module to the cloud and subsequently retrieve the locations
    of other nearby vehicles.
    Integrating the locations of nearby vehicles via the cloud allows the
    planning module to generate safer trajectories which avoid collisions with
    vehicles located in blind spots.

\myparagraph{Contingency planning.} During the course of computation, %
    \av{}s must make probabilistic decisions which affect the outputs of the
    modules.
    For example, object detectors are configured with a confidence threshold
    to filter out misdetections from the model's outputs~\cite{detr-huggingface}.
    Similarly, the prediction module generates several possible
    trajectories for nearby obstacles, and ranks them based on their
    probability of occurring~\cite{rhinehart2018r2p2,rhinehart2019precog}.
    The planning module then uses the most likely trajectory of the
    obstacles to plan a trajectory for an \av{} to follow.
    However, \av{}s can offload the computation of plans that
    handle unlikely object trajectories to the cloud.
    When an object takes an unlikely trajectory, \av{}s can access
    the corresponding cloud-computed plan, enabling quick reactions
    to any sudden changes in the environment.

\cref{s:evaluation} evaluates the efficacy of these mechanisms to ensure the
safety of an \av{} under real-world complex scenarios.

\section{Related Work}
\label{s:related-work}

Our principle design objective, as per \cref{s:techniques}, is to use the
cloud as a best-effort pool of resources in order to improve
the accuracy of \av{} decision-making while retaining the ability to fall
back to locally-computed results to handle delays and variability.
We now discuss the feasibility of implementing such an approach using existing
execution systems.

The Robot Operating System (ROS)~\cite{ros} is the current execution 
system-of-choice for \av{}s and has been deployed by vendors including
Autoware~\cite{autoware}, Cruise~\cite{cruise-roscon}, BMW~\cite{bmw-ros},
and others~\cite{daimler-ros,udacity}.
ROS' highly-modular design enables various initiatives to provide 
additional feature enhancements, such as real-time support~\cite{apexai}).
One such initiative which aims to add cloud support to ROS is
FogROS~\cite{fogros,ichnowski2022fogros}.
FogROS allows users to designate ROS nodes that must execute in the cloud and
specify their resource requirements.
FogROS then provisions suitable cloud instances, configures the network connection
between 
the cloud and the local machine, and registers the cloud nodes with the local
coordinator, enabling seamless communication between the cloud and local nodes.

While FogROS greatly simplifies the deployment of specific ROS nodes to the
cloud, it lacks out-of-the-box support for fall-back mechanisms that
switch to local computation
when the connection to the cloud becomes too unreliable.
Moreover, implementing such an approach atop ROS' callback-execution model
is a complex, tedious, and error-prone process because developers must implement and 
manage varying deadlines (see \cref{s:background}) using
fine-grained wall-clock timers on every node~\cite{ros2-timer}.
Because these fine-grained timers may cause priority inversions when they
rapidly produce events which overfill a \texttt{SingleThreadedExecutor}'s queue,
developers must implement ROS nodes using \texttt{MultiThreadedExecutor}s
which requires manually managing shared state using locks in order to
ensure the safe and correct fusion of results from the cloud and
from local computation.
This complexity further increases if nodes attempt to maximize the accuracy
of computation under a deadline by concurrently executing a mixture of algorithm
and model implementations.

In addition to ROS, several related works have proposed designs that allow
\av{}s to utilize the cloud, however, these approaches focus on specific
tasks such as merging sensor data and intermediate representations across a fleet of 
vehicles~\cite{kumar2012cloud,qiu2018avr,chen2019f,chinchali2019neural,zhang2021emp}.
For example, both Carcel~\cite{kumar2012cloud} and EMP~\cite{zhang2021emp} 
enable a fleet of vehicles to create an accurate representation of their
environment by selectively sharing their data to an edge server while remaining
resilient to network fluctuations.
Other work (e.g., Neurosurgeon~\cite{kang2017neurosurgeon}) has examined
\emph{how} to split computation between the cloud and the 
\av{}~\cite{cui2020offloading,sun2018learning,sun2018cooperative}.
However, none of these works propose a general system design for \av{}s
that falls back to results computed by on-board hardware in order to handle 
faults in the cloud or the network.
Therefore, we believe that these works are complementary,
and that \av{}s can integrate their methods with our design in order to
gain further accuracy improvements while maintaining a reliable baseline
accuracy through the ability to fall back to results from local computation.

\section{Design}
\label{s:design}

Given the limitations of the current systems, we seek to design an execution
system that enables \av{} developers to easily achieve our design goal of 
augmenting the computation's accuracy with the vast pool of resources in the 
cloud.
To ensure that the programming model remains familiar to the current \av{} 
systems (e.g., ROS), our system models an \av{} pipeline as a directed graph
in which vertices, referred to as \textit{operators} are akin to ROS nodes, and
are connected to other vertices via \textit{streams}.
The streams are statically-typed and enforce an interface between the sender
(akin to a ROS publisher) and a receiver (akin to a ROS sender).

\begin{figure}[t!]
  \medskip
    \begin{center}
        \includegraphics[width=0.9\columnwidth]{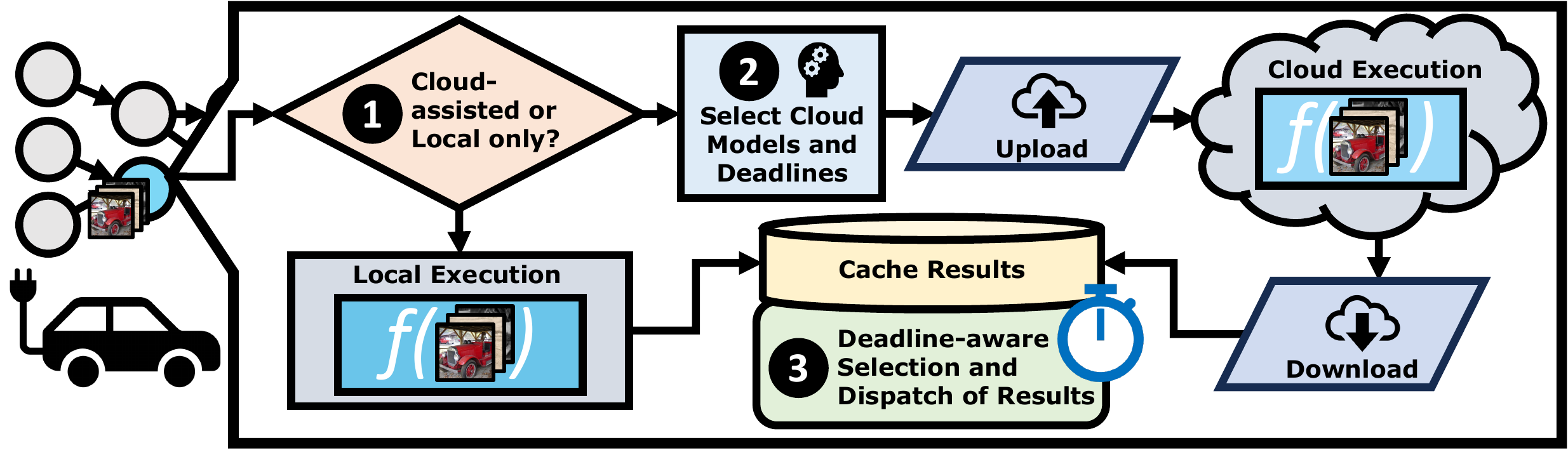}
    \end{center}
    \caption{\textbf{Our speculative cloud execution} design allows each
    operator to make fine-grained decisions about when to contact the cloud
    and how much time to allocate for a response. The results from the cloud
    are automatically incorporated to ensure the highest-possible accuracy.}
    \label{fig:design-diagram}
    \vspace{-1.6em}
\end{figure}

However, to safely exploit the cloud resources in the presence of network
delays and unavailability of cloud resources, our system allows operators to
define \textit{deadlines} that bound the time which can elapse between the 
transmission of a particular request to the cloud and the retrieval of its 
result.
To do so, the system must enable each operator to: \one{} specify the 
computation that executes locally and the computation that can execute in the
cloud, \two{} control what input data is transmitted to the cloud and what 
deadline is assigned to its completion, and \three{} fuse the inputs from the
cloud and the on-board hardware to ensure that the maximum accuracy results
available by the deadline are used.
We emphasize that our system design does not automatically decide which
models and algorithms are suitable for cloud execution, but rather provides
the abstractions and mechanisms to enable augmenting local computation
with the cloud in order to benefit overall accuracy.
\cref{fig:design-diagram} provides an overview of our approach, and we 
now elaborate on how our system achieves each of the aforementioned aspects 
below:

\myparagraph{\bubble{1} Operator configuration.} Each operator must
implement a \texttt{setup} method in which the operator informs the system
if its computation is to be assisted by a cloud implementation.
To do so, the operator initiates a connection to the cloud for dispatching
remote procedure calls (RPCs)
and invokes the \texttt{use\_cloud} method in our system with the following
API:
\vspace{-0.4em}
\begin{center}
    \verb+use_cloud(handle, type, msg_handler, priority)+
\end{center}
\vspace{-0.4em}

The operator registers the \texttt{handle} to the RPC connection with the 
system along with the \texttt{type} of the message to be relayed.
Moreover, the system allows an operator to invoke \texttt{use\_cloud}
multiple times in order to take advantage of multiple cloud
implementations which may span different cloud providers and exploit
vendor-specific hardware to provide the highest-accuracy results within
the given deadline~\cite{stoica2021cloud}.
By setting the \texttt{priority}, the system ensures a total order over
the results of the invocations in order to select which of the potential
results to output.
For example, an object detection operator may invoke several models
from \cref{t:model-detection} in the cloud with the hope that one
of them returns a result by the given deadline, and may set their
priorities according to each detector's accuracy on a static dataset in order
to favor results from more accurate detectors.
Invoking this API informs the underlying execution system that it needs to 
set up the mechanisms detailed below, which ensure the correct collation of
results from the cloud and the local execution.

\myparagraph{\bubble{2} Deadline enforcement.} While the execution of non-cloud
enabled operators continues as normal, cloud-enabled operators must be able to
specify a deadline on the time that they are able to wait for a higher-accuracy
result from one of the cloud implementations registered with
\texttt{use\_cloud}.
To do so, each invocation of \texttt{use\_cloud} requires a
\texttt{msg\_handler} function that conforms to the following API:
     \verb+(input_message, timestamp) ->+ \\
     \verb+           Optional[output_message, deadline]+

For each message that the operator receives, the system invokes the
registered \texttt{msg\_handler}s which may use the message content
and the
timestamp to decide whether to send the message to the 
handler's cloud implementation and what deadline to assign to
this request.
Unlike ROS timestamps that correspond to physical time, this 
\texttt{timestamp} is an internal logical representation of time that 
corresponds to a counter of the messages received by the operator.
In addition to deciding whether to offload the computation of the messages
to the cloud based on its 
contents~\cite{context-aware-streaming,chameleon}, the handler 
functions can use the logical timestamps to enforce simple policies such as 
sending non-consecutive inputs to the cloud to reduce bandwidth usage, or
alternating between different implementations to %
avoid bottlenecks in the cloud service and ensure fault-tolerance for the
operator.

The operator makes the system aware of its intent to transmit a message to the
cloud service by returning an \texttt{output\_message} from the handler.
The \texttt{output\_message} must match the message definition in the RPC handle registered
with \texttt{use\_cloud}, but can be different from the type of the 
\texttt{input\_message}.
This type differentiation allows the operators to bundle extra state to the 
cloud platform that is not communicated to other operators.
For example, the object tracking operator in the perception module assigns
unique identifiers to detected obstacles based on their past history.
As a result, selectively executing inputs to this operator in the
cloud also requires transmitting the state of the identifiers generated by the local
object tracker to allow the cloud-based object tracker to correctly assign identifiers.

In addition to the message to be conveyed to the cloud service, the handler
decides the \textit{relative} \texttt{deadline} $d$ to be assigned to the
arrival of its result.
The ability to return a new deadline $d$ for each incoming message enables the
system to adjust to varying deadlines required by the \av{} to respond to 
dynamicity in the %
environment~\cite{streaming-perception,context-aware-streaming,erdos}.
Given the deadline and the message, the system decides if the request should
be sent based on the length of its outgoing queues, which fill up when
network's quality of service degrades.
If the system elects to send the request, the system indexes the request by its
\texttt{timestamp} and \texttt{priority} and installs a timer that triggers
$d$ milliseconds from the current time.

\myparagraph{\bubble{3} High-accuracy results.} In parallel to executing the 
request in the cloud, the system invokes the callbacks registered for the 
incoming message and executes them on the local hardware with the lowest
\texttt{priority} $p$.
Upon completion, the local callbacks invoke \texttt{send} with the output
message to convey the results to downstream operators.
However, the system intercepts the invocation of the \texttt{send} to check
for pending cloud requests with the same \texttt{timestamp} and a higher
priority $p' > p$.
If a pending cloud request is found, the system caches the local result
and waits for the request's timer to interrupt as detailed below.
However, if no pending cloud request is found and the system already forwarded
a message from a cloud execution to the downstream operators (i.e., the
result from the cloud arrives before the local computation finishes),
the local result is dropped.
Otherwise, the message containing the local result is sent to the downstream operators.

Upon arrival of a message from the cloud with \texttt{priority} $p$ and
\texttt{timestamp} $t$, the system checks if the message missed its deadline by
querying the status of the timer installed for ($t$, $p$).
If the timer was triggered, the message missed its deadline and is dropped.
Similarly, if the system already sent a message for $t$ %
(i.e., a higher priority request returned more quickly or another request's
timer triggered earlier), the message is dropped and the
installed timer is deactivated.
Otherwise, if no higher priority cloud request is pending and the system
has not sent a message for $t$, %
the message is sent to the downstream operators and the installed timer
is deactivated.
However, if there is a pending cloud request with a higher priority, the current
results are cached.

Finally, if an active timer triggers,
the system receives an interrupt in which
it checks its cache for the highest priority message presently available for
the \texttt{timestamp}
and forwards the message to the downstream operators.

\section{Evaluation}
\label{s:evaluation}
We now evaluate the efficacy of our approach in augmenting the safety of 
cloud-assisted \av{}s under complex, real-world scenarios while maintaining 
the accuracy of an \av{} using only on-board computation.
We evaluate our design to demonstrate that our approach to augmenting \av{}s
Specifically, we seek to answer the following questions: 
\begin{enumerate}
  \item Do current technologies support a low-latency and reliable connection 
      to the cloud? (\cref{s:evaluation:feasibility})
  \item Does our system enable the three techniques (\cref{s:techniques}) to
      exploit cloud resources to enhance \av{} safety? 
        (\cref{s:evaluation:scenarios})
\end{enumerate}

\begin{figure}[t!]
    \medskip
    \begin{center}
      \includegraphics[width=0.8\columnwidth]{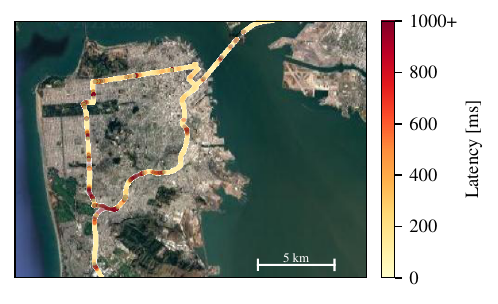}
    \end{center}
    \vspace{-1.3em}
    \caption{\textbf{Cellular network latency} of a 5G connection while
    driving through a route in San Francisco frequented by Waymo and Cruise
    \av{}s.}
    \label{f:sf-map-drive}
\end{figure}

\subsection{Feasibility of Cloud Access}
\label{s:evaluation:feasibility}
We investigate whether modern cellular networks are able to provide the speed
and bandwidth necessary to execute data-intensive \av{} operators.
To measure network performance in realistic setting,
we conduct a field test by following a route in San Francisco where Cruise and 
Waymo already provide fully autonomous rides~\cite{cruise-welcome-riders, 
waymo-sf-riders}.
The route contains both urban and highway driving and is visualized in 
\cref{f:sf-map-drive}. 
We discuss our experiment setup and the findings below.

\myparagraph{Experiment Setup.}
We model the transmission of HD camera footage from an \av{} to the cloud.
In our vehicle, we connect a Lenovo ThinkPad P1 Gen 2 laptop to an
Inseego MiFi X PRO 5G hotspot on the Verizon network via USB-C.
On a Google Pixel 5, we collect timestamped GPS coordinates at $1$ Hz.
The laptop executes a multithreaded gRPC~\cite{grpc} client which sends $33.3$ KB messages
at $30$ Hz to a server to match the bitrate of 30 FPS HD camera
footage~\cite{youtube-bitrate}.
We establish a connection to a Google Cloud Platform \texttt{n1-highmem-8}
instance in the \texttt{us-west2-a} zone which executes a gRPC server
that responds to messages from the client with $1$ KB acknowledgments.
The client measures the round-trip latency of sending a message to receiving
an reply. %
Because network delays may cause the client to queue messages and overstate
the network latency,
the client waits for all messages to process if it detects more than 30
queued messages.

\myparagraph{Findings.}
The 5G network in San Francisco frequently provides latencies that
enable cloud execution.
We measure a median round-trip-latency of $68$ ms (\cref{f:sf-map-drive})
which demonstrates an opportunity
to take advantage of the hundreds of milliseconds in runtime disparity between
cloud and \av{} hardware (\cref{t:model-detection}).
In addition, the long tail of network latencies from $336$ ms at the $90$th percentile
to $3027$ ms at the $99$th percentile substantiates the need to %
manage network delays.
In \cref{s:evaluation:scenarios}, we demonstrate that our design takes advantage
of the
fast common-case latencies to enable critical safety benefits while mitigating
the impact of long tail-end latencies by falling back to results from local
computation.

\subsection{Study of Scenarios}
\label{s:evaluation:scenarios}

We study of the safety benefits of our design (\cref{s:design}) and how it
enables the three mechanisms which use the cloud to improve \av{} safety (\cref{s:techniques}).
For each mechanism, we demonstrate its efficacy under a complex, real-world scenario
executed %
using the
CARLA simulator~\cite{carla}.
We use the pseudo-asynchronous mode of execution from the Pylot \av{}
platform~\cite{gog2021pylot} to simulate the delay of calculating a
demand for different end-to-end deadlines, retrieved from
different end-to-end deadlines, retrieved from
\cref{f:sf-map-drive}\footnote{Videos of scenarios are available at
\href{https://tinyurl.com/26fzrabu}{https://tinyurl.com/26fzrabu}}.

\begin{figure}[t!]
    \medskip
    \begin{center}
        \includegraphics[width=0.9\columnwidth]{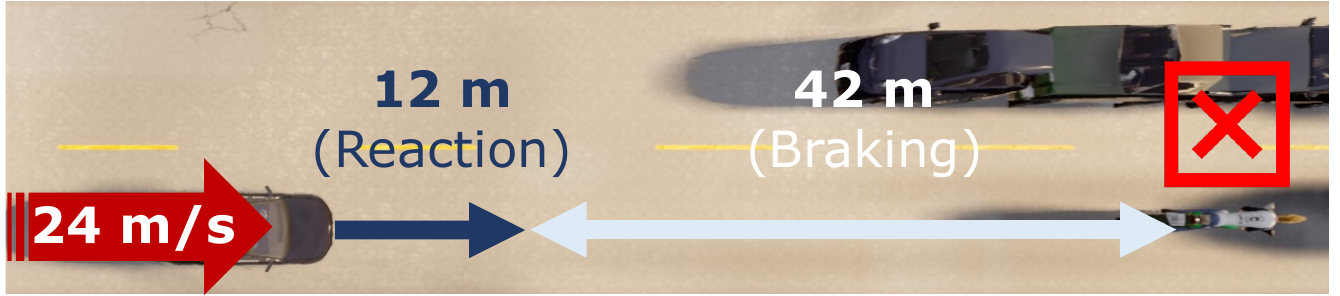}
        \includegraphics[width=0.9\columnwidth]{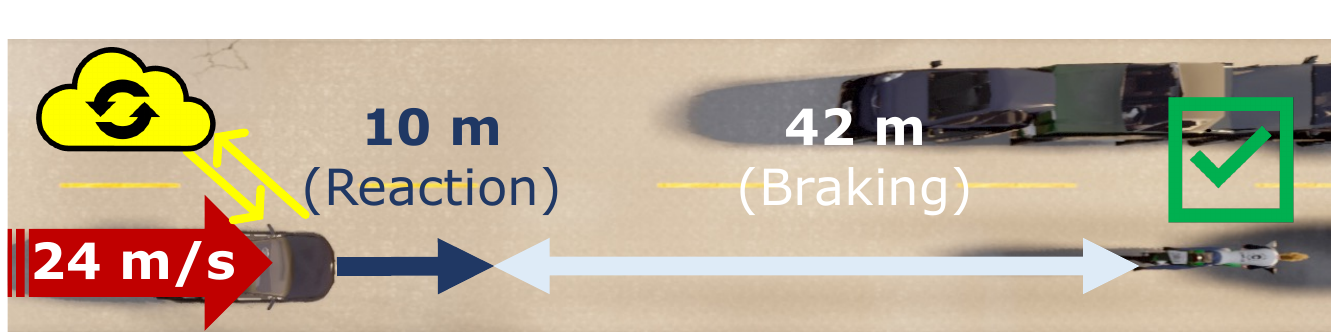}
    \end{center}
    \caption{\textbf{Traffic Jam Scenario} leverages the cloud's ability to 
    run \textit{higher-accuracy models} at a reduced latency to reduce the
    \av{}'s response time, thus minimizing its reaction time and avoiding a 
    collision with the motorcycle.}
    \label{f:traffic-jam}
\end{figure}

\begin{figure}[t!]
    \centering
    \begin{subfigure}[b]{0.23\textwidth}
        \centering
        \includegraphics[width=\textwidth]{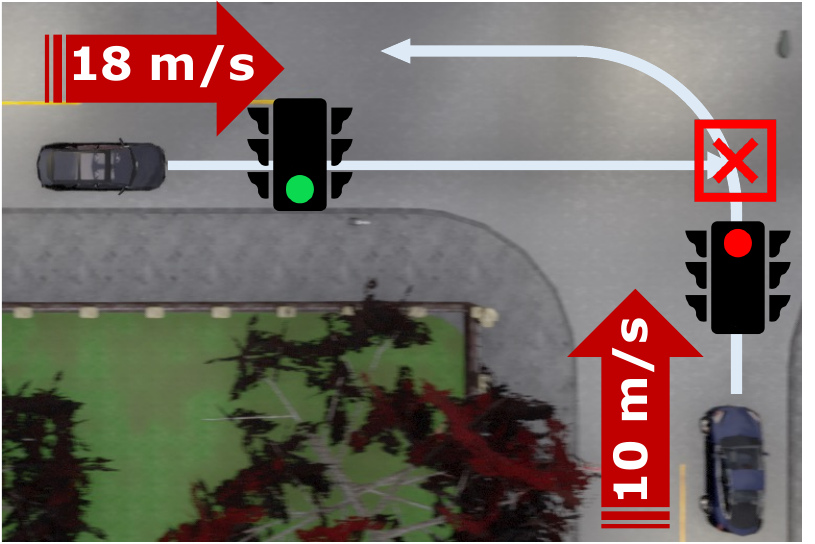}
    \end{subfigure}
    \begin{subfigure}[b]{0.23\textwidth}
        \centering
        \includegraphics[width=\textwidth]{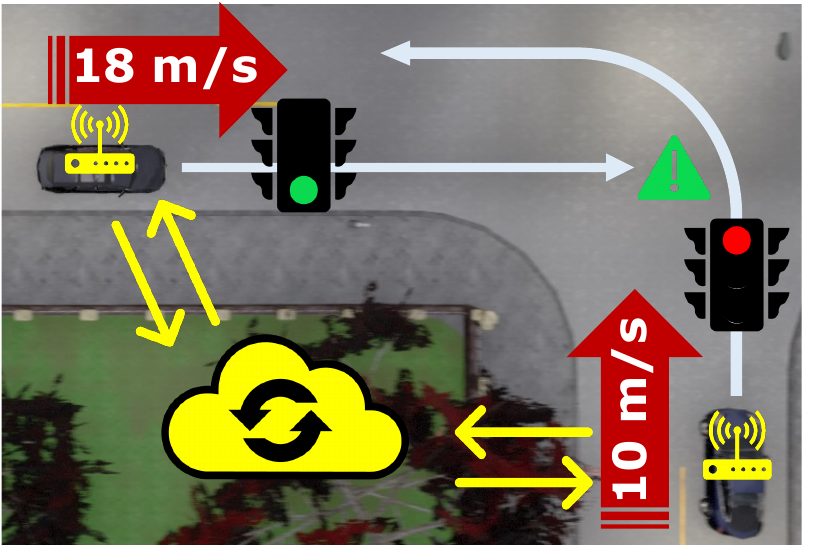}
    \end{subfigure}
    \caption{\textbf{Running a Red Light Scenario} leverages the cloud's
    ability to build \textit{accurate environment representations} to detect
    the occluded vehicle running the red light in time to avoid a collision.}
    \label{f:intersection}
\end{figure}

\myparagraph{Traffic Jam.} This scenario from \cite{erdos}
simulates merging into a traffic jam, visualized in \cref{f:traffic-jam}.
The \av{} drives at a high speed on a two lane undivided road and must come to a halt
behind the motorcycle stopped in the distance.
The motorcycle complicates this scenario because the \av{} must perceive the
stopped obstacle from afar to prevent a collision.
Moreover, the \av{} cannot swerve to avoid a collision due the vehicles in the
opposite lane.
Since this scenario requires both far-away detections and rapid responses
due to the high driving speed, the technique of exploiting the cloud to
run \textit{higher-accuracy models} quickly ensures maximum safety.

To evaluate safety, we use a simple planner that brakes
once the \av{}'s object detection operator
identifies the obstacle on three consecutive camera frames.
We then investigate the following three operator configurations
(\cref{t:traffic-jam-arbitrage}):

\begin{mylist}
    \item{} \textbf{Local} which executes DETR-ResNet-$50$ on a local NVIDIA 
        Orin GPU. We choose DETR-ResNet-50 since it is the only model that 
        provides response times required to ensure human-level safety on the
        local GPU (\cref{t:model-detection} and \cref{s:introduction}).
    \item{} \textbf{Cloud} which executes DETR-ResNet-$101$ on an NVIDIA 
        A$100$ GPU running on a Google Cloud \texttt{a2-highgpu-1g} instance. 
        We choose DETR-ResNet-$101$ to ensure that the local and
        cloud models belong to the same architecture.
    \item{} \textbf{Ours} which enables operators to specify deadlines on
        response time from the cloud and fall back to local results when the
        cloud is unable to meet the deadline.
\end{mylist}

We sweep the entire range of driving speeds in California (i.e., $25$ mph to 
$65$ mph), and simulate cloud response times up to the p$99$ latency collected from 
our drive through San Francisco (\cref{s:evaluation:feasibility}).
We note that the higher cloud response times do not apply to the \textit{Local} 
approach.
We find that at higher speeds (e.g., $22$ m/s), the lower-latency access 
to \textit{higher-accuracy models} afforded by the cloud is critical in 
ensuring the safety of the \av{}.
However, without appropriate mechanisms to fall back to local computation using
deadlines as proposed in our design, the \textit{Cloud} approach incurs more
safety violations than the \textit{Local} approach when network latency is
high (e.g., a $3$ second cloud response time when the \av{} drives at $18$ m/s).

\cref{f:traffic-jam} investigates how executing a higher-accuracy object
detector locally affects collision-avoidance in high-speed scenarios.
We compare a \textit{Local} execution of DETR-ResNet-$101$ to a
\textit{Cloud} execution with median network latency.
We find that the local execution fails to break in time due to its
$222$ ms longer response time, resulting in a collision.

\myparagraph{Running a Red Light.}
This scenario from the NHTSA pre-crash 
typology~\cite{carla-nhtsa-scenarios, nhtsa-precrash-scenarios}
simulates a vehicle running a red light
which forces the \av{} to perform a collision avoidance maneuver.
The \av{}s can only perceive the other vehicle just before a potential collision,
challenging its ability to respond in time.
We evaluate three variations of this scenario with different intersections,
and find that neither local nor cloud executions of object detectors are
able to avoid a collision.

However, using the cloud's ability to build an \textit{accurate environment
representation} (e.g., by sharing location data with nearby \av{}s),
allows the \av{} to plan its trajectory with other vehicles in its blind spots.
This enables the \av{} to brake early and avoid a collision in this scenario.

\begin{table}[t!]
\medskip
\scriptsize
\caption{Configurations that avoid a collision are marked in 
    \textcolor{green}{green}, while configurations that collide are marked in
    \textcolor{red}{red}.}
    \vspace{-1.0em}
\centering
  \setlength\tabcolsep{1.5pt}
\begin{tabularx}{\columnwidth}{c!{\vrule width 1.5pt}c!{\vrule width 1.5pt}X|X|X|X|X|X}
  \textbf{Speed} & \multirow{2}{*}{\textbf{Approach}}
    & \multicolumn{6}{|c}{\textbf{Cloud Response Time [s]}} \\ \hhline{~|~|-|-|-|-|-|-} 
    \textbf{[m/s]} & & 0.5 & 0.75 & 1.0 & 1.25 & 1.5 & 3.0 \\
  \noalign{\hrule height 1.5pt}
  \multirow{3}{*}{11} & \textbf{Local}
    & \cellcolor{green} & \cellcolor{green} & \cellcolor{green}
    & \cellcolor{green} & \cellcolor{green} & \cellcolor{green}
    \\ \hhline{~|-|-|-|-|-|-|-}
  & \textbf{Cloud}
    & \cellcolor{green} & \cellcolor{green} & \cellcolor{green}
    & \cellcolor{green} & \cellcolor{green} & \cellcolor{green}
    \\ \hhline{~|-|-|-|-|-|-|-}
  & \textbf{Ours}
    & \cellcolor{green} & \cellcolor{green} & \cellcolor{green}
    & \cellcolor{green} & \cellcolor{green} & \cellcolor{green}
  \\ \noalign{\hrule height 1.5pt}
  \multirow{3}{*}{18} & \textbf{Local}
    & \cellcolor{green} & \cellcolor{green} & \cellcolor{green}
    & \cellcolor{green} & \cellcolor{green} & \cellcolor{green}
    \\ \hhline{~|-|-|-|-|-|-|-}
  & Cloud
    & \cellcolor{green} & \cellcolor{green} & \cellcolor{green}
    & \cellcolor{green} & \cellcolor{green} & \cellcolor{red}
    \\ \hhline{~|-|-|-|-|-|-|-}
  & \textbf{Ours}
    & \cellcolor{green} & \cellcolor{green} & \cellcolor{green}
    & \cellcolor{green} & \cellcolor{green} & \cellcolor{green}
  \\ \noalign{\hrule height 1.5pt}
  \multirow{3}{*}{20} & \textbf{Local}
    & \cellcolor{green} & \cellcolor{green} & \cellcolor{green}
    & \cellcolor{green} & \cellcolor{green} & \cellcolor{green}
    \\ \hhline{~|-|-|-|-|-|-|-}
  & Cloud
    & \cellcolor{green} & \cellcolor{green} & \cellcolor{green}
    & \cellcolor{green} & \cellcolor{red} & \cellcolor{red}
    \\ \hhline{~|-|-|-|-|-|-|-}
  & \textbf{Ours}
    & \cellcolor{green} & \cellcolor{green} & \cellcolor{green}
    & \cellcolor{green} & \cellcolor{green} & \cellcolor{green}
  \\ \noalign{\hrule height 1.5pt}
 \multirow{3}{*}{22} & Local
    & \cellcolor{red} & \cellcolor{red} & \cellcolor{red}
    & \cellcolor{red} & \cellcolor{red} & \cellcolor{red}
    \\ \hhline{~|-|-|-|-|-|-|-}
  & \textbf{Cloud}
    & \cellcolor{green} & \cellcolor{green} & \cellcolor{red}
    & \cellcolor{red} & \cellcolor{red} & \cellcolor{red}
    \\ \hhline{~|-|-|-|-|-|-|-}
  & \textbf{Ours}
    & \cellcolor{green} & \cellcolor{green} & \cellcolor{red}
    & \cellcolor{red} & \cellcolor{red} & \cellcolor{red}
  \\ \noalign{\hrule height 1.5pt}
 \multirow{3}{*}{24} & Local
    & \cellcolor{red} & \cellcolor{red} & \cellcolor{red}
    & \cellcolor{red} & \cellcolor{red} & \cellcolor{red}
    \\ \hhline{~|-|-|-|-|-|-|-}
  & \textbf{Cloud}
    & \cellcolor{green} & \cellcolor{red} & \cellcolor{red}
    & \cellcolor{red} & \cellcolor{red} & \cellcolor{red}
    \\ \hhline{~|-|-|-|-|-|-|-}
  & \textbf{Ours}
    & \cellcolor{green} & \cellcolor{red} & \cellcolor{red}
    & \cellcolor{red} & \cellcolor{red} & \cellcolor{red}
\end{tabularx}

\label{t:traffic-jam-arbitrage}
\centering
    \vspace{-1.6em}
\end{table}

\begin{figure}[t!]
    \medskip
    \begin{center}
        \includegraphics[width=0.9\columnwidth]{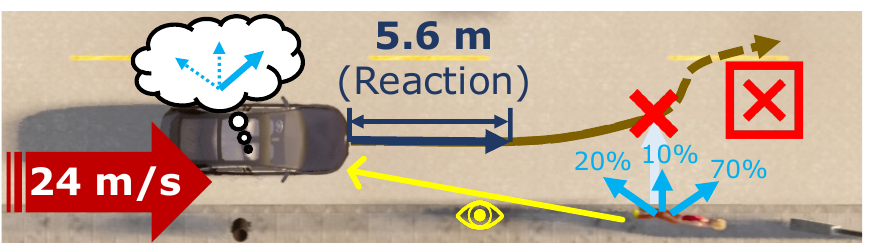}
        \includegraphics[width=0.9\columnwidth]{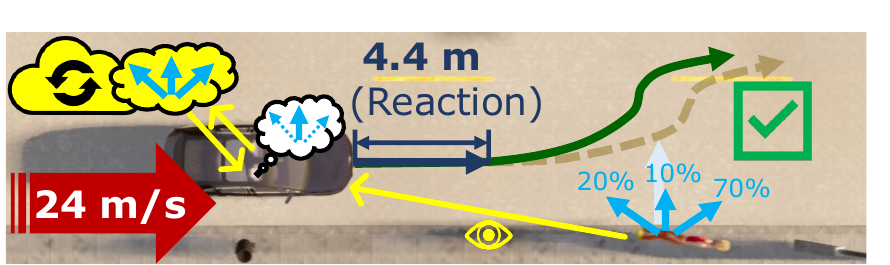}
    \end{center}
    \caption{\textbf{Person Jaywalking Scenario} leverages the cloud's
    ability to do \textit{contingency planning} 
    for the unlikely case that the pedestrian enters the street, allowing the
    \av{} to use the cached plan quickly and avoid a collision.
    }
    \label{f:contingency-planning}
     \vspace{-1.3em}
\end{figure}

\myparagraph{Person Jaywalking.}
This scenario simulates a person unexpectedly entering the street, requiring
the \av{} to quickly respond in order to avoid a collision.
\cref{f:contingency-planning} evaluates the scenario in which an \av{} executes
its planning module locally, which has a $500$ ms end-to-end response time.
Because the pedestrian enters the street when the \av{} is only $10$ m away,
the \av{} cannot generate an emergency swerving maneuver or stop in time, resulting
in a collision with the pedestrian.

However, we perform \textit{contingency planning} using the cloud which
generates a plan for the low-likelihood case that the pedestrian enters the street,
downloads the plan to the \av{}, and caches the plan in the \av{}'s planner.
When the cloud-assisted \av{} detects the pedestrian entering the street, the \av{} 
enacts the cached contingency plan and bypasses the local planner, lowering 
the response time to $400$ ms.
We observe that the cloud-computed contingency plan enables the \av{}
to swerve in time and successfully avoid a collision.

\section{Discussion \& Conclusion}
\label{s:discussion}
Our experiments show that \av{}s can leverage the cloud with current
technology to exploit safety benefits.
We present a design which ensures that cloud execution
strictly improves safety by executing models and algorithms
\textit{speculatively} in the cloud.
When the cloud fails to produce results by a deadline,
our design falls back to results from local execution which
are computed concurrently.
Although our design focuses on the enforcement of provided deadlines,
the calculation of deadlines for cloud execution presents an
exciting opportunity for further exploration.

Moreover, trends in technology demonstrate that \av{}s of the future
will harness the benefits of the cloud.
\av{}s increasingly rely on higher-fidelity data collected from more
sensors which they process using larger and more compute-intensive
algorithms and models.
Meanwhile, cloud and network technologies continue to improve, and
will present more arbitrage opportunities in which the cloud can
benefit safety.
We hope that this research will bring attention to the potential
advantages of augmenting \av{}s with the cloud and inspire further
exploration and development in this area.

{

  \balance
  \bibliographystyle{IEEEtran}
  \bibliography{paper}

\begin{thebibliography}{10}
\providecommand{\url}[1]{#1}
\csname url@samestyle\endcsname
\providecommand{\newblock}{\relax}
\providecommand{\bibinfo}[2]{#2}
\providecommand{\BIBentrySTDinterwordspacing}{\spaceskip=0pt\relax}
\providecommand{\BIBentryALTinterwordstretchfactor}{4}
\providecommand{\BIBentryALTinterwordspacing}{\spaceskip=\fontdimen2\font plus
\BIBentryALTinterwordstretchfactor\fontdimen3\font minus
  \fontdimen4\font\relax}
\providecommand{\BIBforeignlanguage}[2]{{%
\expandafter\ifx\csname l@#1\endcsname\relax
\typeout{** WARNING: IEEEtran.bst: No hyphenation pattern has been}%
\typeout{** loaded for the language `#1'. Using the pattern for}%
\typeout{** the default language instead.}%
\else
\language=\csname l@#1\endcsname
\fi
#2}}
\providecommand{\BIBdecl}{\relax}
\BIBdecl

\bibitem{national-high-safety-2020}
{National Highway Traffic Safety Administration}, ``{Traffic Safety Facts (2020
  Data)},''
  \url{https://crashstats.nhtsa.dot.gov/Api/Public/ViewPublication/813375}.

\bibitem{erdos}
I.~Gog, S.~Kalra, P.~Schafhalter, J.~E. Gonzalez, and I.~Stoica, ``D3: {A}
  {D}ynamic {D}eadline-{D}riven approach for {B}uilding {A}utonomous
  {V}ehicles,'' in \emph{Proceedings of the Seventeenth European Conference on
  Computer Systems}, 2022, pp. 453--471.

\bibitem{wang2020safety}
J.~Wang, L.~Zhang, Y.~Huang, J.~Zhao, and F.~Bella, ``Safety of autonomous
  vehicles,'' \emph{Journal of advanced transportation}, vol. 2020, pp. 1--13,
  2020.

\bibitem{gog2021pylot}
I.~Gog, S.~Kalra, P.~Schafhalter, M.~A. Wright, J.~E. Gonzalez, and I.~Stoica,
  ``{Pylot: A Modular Platform for Exploring Latency-Accuracy Tradeoffs in
  Autonomous Vehicles},'' in \emph{Proceedings of the IEEE International
  Conference on Robotics and Automation (ICRA)}.\hskip 1em plus 0.5em minus
  0.4em\relax IEEE, 2021, pp. 8806--8813.

\bibitem{wang2022adept}
S.~Wang, Z.~Sheng, J.~Xu, T.~Chen, J.~Zhu, S.~Zhang, Y.~Yao, and X.~Ma,
  ``{ADEPT}: A testing platform for simulated autonomous driving,'' in
  \emph{37th IEEE/ACM International Conference on Automated Software
  Engineering}, 2022, pp. 1--4.

\bibitem{dreossi2019verifai}
T.~Dreossi, D.~J. Fremont, S.~Ghosh, E.~Kim, H.~Ravanbakhsh,
  M.~Vazquez-Chanlatte, and S.~A. Seshia, ``Verif{AI}: A toolkit for the formal
  design and analysis of artificial intelligence-based systems,'' in
  \emph{Computer Aided Verification: 31st International Conference, CAV 2019,
  New York City, NY, USA, July 15-18, 2019, Proceedings, Part I 31}.\hskip 1em
  plus 0.5em minus 0.4em\relax Springer, 2019, pp. 432--442.

\bibitem{tesla-hardware-4}
``Tesla {FSD} {H}ardware 4.0 {R}evealed: {M}ore {C}ameras, {N}ew
  {P}lacements,'' \url{https://tinyurl.com/yc2as3f6}.

\bibitem{waymo-4th-gen-specs}
``Waymo one autonomous robotaxi first ride: Way mo' better than driving?''
  \url{https://tinyurl.com/yvku75kx}.

\bibitem{waymo-av-sensors}
``{Introducing the 5\textsuperscript{th} Generation Waymo Driver},''
  \url{https://blog.waymo.com/2020/03/introducing-5th-generation-waymo-driver.html}.

\bibitem{liu2021swin}
Z.~Liu, Y.~Lin, Y.~Cao, H.~Hu, Y.~Wei, Z.~Zhang, S.~Lin, and B.~Guo, ``Swin
  transformer: Hierarchical vision transformer using shifted windows,'' in
  \emph{Proceedings of the IEEE/CVF international conference on computer
  vision}, 2021, pp. 10\,012--10\,022.

\bibitem{li2022exploring}
Y.~Li, H.~Mao, R.~Girshick, and K.~He, ``Exploring plain vision transformer
  backbones for object detection,'' in \emph{Computer Vision--ECCV 2022: 17th
  European Conference, Tel Aviv, Israel, October 23--27, 2022, Proceedings,
  Part IX}.\hskip 1em plus 0.5em minus 0.4em\relax Springer, 2022, pp.
  280--296.

\bibitem{liu2022convnet}
Z.~Liu, H.~Mao, C.-Y. Wu, C.~Feichtenhofer, T.~Darrell, and S.~Xie, ``A convnet
  for the 2020s,'' in \emph{Proceedings of the IEEE/CVF Conference on Computer
  Vision and Pattern Recognition}, 2022, pp. 11\,976--11\,986.

\bibitem{zhai2022scaling}
X.~Zhai, A.~Kolesnikov, N.~Houlsby, and L.~Beyer, ``Scaling vision
  transformers,'' in \emph{Proceedings of the IEEE/CVF Conference on Computer
  Vision and Pattern Recognition}, 2022, pp. 12\,104--12\,113.

\bibitem{tan2021efficientnetv2}
M.~Tan and Q.~Le, ``Efficientnetv2: Smaller models and faster training,'' in
  \emph{International conference on machine learning}.\hskip 1em plus 0.5em
  minus 0.4em\relax PMLR, 2021, pp. 10\,096--10\,106.

\bibitem{tan20efficientdet}
M.~Tan, R.~Pang, and Q.~V. Le, ``{EfficientDet: Scalable and Efficient Object
  Detection},'' in \emph{Proceedings of the IEEE Conference on Computer Vision
  and Pattern Recognition (CVPR)}, 2020.

\bibitem{driving-michigan}
\BIBentryALTinterwordspacing
S.-C. Lin, Y.~Zhang, C.-H. Hsu, M.~Skach, M.~E. Haque, L.~Tang, and J.~Mars,
  ``{The Architectural Implications of Autonomous Driving: Constraints and
  Acceleration},'' in \emph{Proceedings of the 23\textsuperscript{rd}
  International Conference on Architectural Support for Programming Languages
  and Operating Systems (ASPLOS)}, 2018, pp. 751--766. [Online]. Available:
  \url{http://doi.acm.org/10.1145/3173162.3173191}
\BIBentrySTDinterwordspacing

\bibitem{wolfe2020rapid}
B.~Wolfe, B.~Seppelt, B.~Mehler, B.~Reimer, and R.~Rosenholtz, ``Rapid holistic
  perception and evasion of road hazards.'' \emph{Journal of experimental
  psychology: general}, vol. 149, no.~3, p. 490, 2020.

\bibitem{johansson1971drivers}
G.~Johansson and K.~Rumar, ``Drivers' brake reaction times,'' \emph{Human
  factors}, vol.~13, no.~1, pp. 23--27, 1971.

\bibitem{nvidia-drive}
``{NVIDIA} {DRIVE}: {H}ardware for {S}elf-{D}riving {C}ars,''
  \url{https://www.nvidia.com/en-us/self-driving-cars/drive-platform/hardware/}.

\bibitem{nvidia-orin}
``{NVIDIA} {I}ntroduces {DRIVE} {AGX} {O}rin,''
  \url{https://tinyurl.com/6pjsxzw7}.

\bibitem{nvidia-drive-atlan}
T.~Tomazin, ``{A} {D}ata {C}enter on {W}heels: {NVIDIA} {U}nveils {DRIVE}
  {A}tlan {A}utonomous {V}ehicle {P}latform,''
  \url{https://blogs.nvidia.com/blog/2021/04/12/nvidia-drive-atlan-autonomous-vehicle-platform/}.

\bibitem{nvidia-orin-hyperion}
``{NVIDIA} {E}nters {P}roduction {W}ith {DRIVE} {O}rin, {U}nveils {N}ext-{G}en
  {DRIVE} {H}yperion {AV} platform,'' \url{https://tinyurl.com/2s38djwr}.

\bibitem{tesla-hardware-retrofit}
``Tesla {T}alks {FSD} {H}ardware 4.0, but {T}here {W}ill {N}ot {B}e
  {R}etrofits,''
  \url{https://www.notateslaapp.com/news/1172/tesla-talks-fsd-hardware-4-0-but-there-will-not-be-retrofits}.

\bibitem{nytimes-pace-mismatch}
``As {A}utomakers {A}dd {T}echnology to {C}ars, {S}oftware {B}ugs {F}ollow,''
  \url{https://www.nytimes.com/2022/02/08/business/car-software-lawsuits.html}.

\bibitem{fox2009above}
A.~Fox, R.~Griffith, A.~Joseph, R.~Katz, A.~Konwinski, G.~Lee, D.~Patterson,
  A.~Rabkin, I.~Stoica \emph{et~al.}, ``Above the clouds: A berkeley view of
  cloud computing,'' \emph{Dept. Electrical Eng. and Comput. Sciences,
  University of California, Berkeley, Rep. UCB/EECS}, vol.~28, 2009.

\bibitem{nvidia-gpu}
``{NVIDIA} {GPU} {C}loud {C}omputing,''
  \url{https://www.nvidia.com/en-us/data-center/gpu-cloud-computing/}.

\bibitem{google-tpu}
``{G}oogle {C}loud {TPU},'' \url{https://cloud.google.com/tpu}.

\bibitem{aws-inferentia}
``{AWS} {I}nferentia,''
  \url{https://aws.amazon.com/machine-learning/inferentia/}.

\bibitem{aws-instances}
``{AWS} {EC}2 instance timeline,'' \url{https://instancetyp.es}.

\bibitem{john-wilkes-talk}
``{B}uilding {W}arehouse-{S}cale {C}omputers at {G}oogle {C}loud,''
  \url{https://www.youtube.com/watch?v=9i7HuU8d3_4}.

\bibitem{tesla-recall}
``{T}esla recalls 362,000 {U}.{S}. vehicles over {F}ull {S}elf-{D}riving
  software,''
  \url{https://www.reuters.com/business/autos-transportation/tesla-recalls-362000-us-vehicles-over-full-self-driving-software-2023-02-16/}.

\bibitem{toyota-recall-14}
``{T}oyota recalls 1.9 million cars for software glitch,''
  \url{https://www.cnbc.com/2014/02/12/toyota-recalls-19-million-cars-for-software-glitch.html}.

\bibitem{cruise-report}
{General Motors}, ``{2018 Self-driving safety report},''
  \url{https://www.gm.com/content/dam/company/docs/us/en/gmcom/gmsafetyreport.pdf}.

\bibitem{ntsb-uber}
``{NTSB's Accident Report on the Uber Self-Driving Vehicle Crash},''
  \url{https://tinyurl.com/y334xnez}.

\bibitem{autoware}
{Autoware}, ``{Autoware User's Manual - Document Version 1.1},''
  \url{https://tinyurl.com/2v2jkk9n}.

\bibitem{apollo-baidu-new}
``Architectural overview of {B}aidu's {A}pollo{A}uto,''
  \url{https://github.com/ApolloAuto/apollo#architecture}.

\bibitem{apollo-tl}
``{Apollo's Traffic Light Perception},''
  \url{https://github.com/ApolloAuto/apollo/blob/master/docs/06_Perception/traffic_light.md}.

\bibitem{apollo-baidu-hw}
``Hardware prerequisites of {B}aidu's {A}pollo{A}uto,''
  \url{https://github.com/ApolloAuto/apollo#prerequisites}.

\bibitem{nvidia-drive-orin-spec}
``{NVIDIA} {D}rive {AGX} {O}rin {D}eveloper {K}it,''
  \url{https://tinyurl.com/u2rxefpt}.

\bibitem{nvidia-a100-spec}
``{NVIDIA} {A}100 {S}pecifications,''
  \url{https://www.nvidia.com/en-us/data-center/a100/}.

\bibitem{carion2020end}
N.~Carion, F.~Massa, G.~Synnaeve, N.~Usunier, A.~Kirillov, and S.~Zagoruyko,
  ``End-to-end object detection with transformers,'' in \emph{Computer
  Vision--ECCV 2020: 16th European Conference, Glasgow, UK, August 23--28,
  2020, Proceedings, Part I 16}.\hskip 1em plus 0.5em minus 0.4em\relax
  Springer, 2020.

\bibitem{detr-huggingface}
``{DETR} - {H}ugging {F}ace,''
  \url{https://huggingface.co/docs/transformers/model_doc/detr}.

\bibitem{zhang2022dino}
H.~Zhang, F.~Li, S.~Liu, L.~Zhang, H.~Su, J.~Zhu, L.~Ni, and H.~Shum, ``Dino:
  Detr with improved denoising anchor boxes for end-to-end object detection,''
  in \emph{International Conference on Learning Representations}, 2022.

\bibitem{fogros}
K.~E. Chen, Y.~Liang, N.~Jha, J.~Ichnowski, M.~Danielczuk, J.~Gonzalez,
  J.~Kubiatowicz, and K.~Goldberg, ``Fogros: An adaptive framework for
  automating fog robotics deployment,'' in \emph{2021 IEEE 17th International
  Conference on Automation Science and Engineering (CASE)}.\hskip 1em plus
  0.5em minus 0.4em\relax IEEE, 2021, pp. 2035--2042.

\bibitem{streaming-perception}
M.~Li, Y.~Wang, and D.~Ramanan, ``{Towards Streaming Perception},'' in
  \emph{Proceedings of the European Conference on Computer Vision (ECCV)}, Aug.
  2020.

\bibitem{context-aware-streaming}
G.-E. Sela, I.~Gog, J.~Wong, K.~K. Agrawal, X.~Mo, S.~Kalra, P.~Schafhalter,
  E.~Leong, X.~Wang, B.~Balaji, J.~E. Gonzalez, and I.~Stoica, ``Context-aware
  streaming perception in dynamic environments,'' in \emph{Proceedings of the
  European Conference on Computer Vision (ECCV)}, 2022.

\bibitem{claussmann2019review}
L.~Claussmann, M.~Revilloud, D.~Gruyer, and S.~Glaser, ``{A Review of Motion
  Planning for Highway Autonomous Driving},'' \emph{IEEE Transactions on
  Intelligent Transportation Systems}, vol.~21, no.~5, pp. 1826--1848, 2019.

\bibitem{alcon2020timing}
M.~Alcon, H.~Tabani, L.~Kosmidis, E.~Mezzetti, J.~Abella, and F.~J. Cazorla,
  ``{Timing of Autonomous Driving Software: Problem Analysis and Prospects for
  Future Solutions},'' in \emph{Proceedings of the 26\textsuperscript{th} IEEE
  Real-Time and Embedded Technology and Applications Symposium (RTAS)}.\hskip
  1em plus 0.5em minus 0.4em\relax IEEE, 2020, pp. 267--280.

\bibitem{kehoe2015survey}
B.~Kehoe, S.~Patil, P.~Abbeel, and K.~Goldberg, ``A survey of research on cloud
  robotics and automation,'' \emph{IEEE Transactions on automation science and
  engineering}, vol.~12, no.~2, pp. 398--409, 2015.

\bibitem{rhinehart2018r2p2}
N.~Rhinehart, K.~M. Kitani, and P.~Vernaza, ``{R2P2: A Reparameterized
  Pushforward Policy for Diverse, Precise Generative Path Forecasting},'' in
  \emph{Proceedings of the European Conference on Computer Vision (ECCV)},
  2018, pp. 772--788.

\bibitem{rhinehart2019precog}
N.~Rhinehart, R.~McAllister, K.~Kitani, and S.~Levine, ``{PRECOG: Prediction
  Conditioned on Goals in Visual Multi-Agent Settings},'' in \emph{Proceedings
  of the IEEE International Conference on Computer Vision (CVPR)}, 2019, pp.
  2821--2830.

\bibitem{ros}
M.~Quigley, K.~Conley, B.~Gerkey, J.~Faust, T.~Foote, J.~Leibs, R.~Wheeler, and
  A.~Y. Ng, ``{ROS: An Open-Source Robot Operating System},'' in
  \emph{Proceedings of the IEEE International Conference on Robotics and
  Automation (ICRA); Workshop on Open Source Robotics}, vol.~3, May 2009, p.~5.

\bibitem{cruise-roscon}
N.~Valigi, ``{Lessons Learned Building a Self-Driving Car on ROS},''
  \url{https://roscon.ros.org/2018/presentations/ROSCon2018_LessonsLearnedSelfDriving.pdf},
  2018.

\bibitem{bmw-ros}
M.~Aeberhard, T.~K{\"u}hbeck, B.~Seidl, M.~Friedl, J.~Thomas, and O.~Scheickl,
  ``{Automated Driving with ROS at BMW},''
  \url{http://www.ros.org/news/2016/05/michael-aeberhard-bmw-automated-driving-with-ros-at-bmw.html}.

\bibitem{daimler-ros}
A.~Fregin, M.~Roth, M.~Braun, S.~Krebs, and F.~Flohr, ``{Building a Computer
  Vision Research Vehicle with ROS},''
  \url{http://www.ros.org/news/2018/07/roscon-2017-building-a-computer-vision-research-vehicle-with-ros----andreas-fregin.html}.

\bibitem{udacity}
{Udacity}, ``{An Open Source Self-Driving Car},''
  \url{https://www.udacity.com/self-driving-car}.

\bibitem{apexai}
C.~Ho, S.~Nirmal, J.~P. Samper, S.~Nikulin, A.~Pemmaiah, D.~Pangercic, and
  J.~Becker, ``{ROS2 on Autonomous Vehicles},''
  \url{https://roscon.ros.org/2018/presentations/ROSCon2018_ROS2onAutonomousDrivingVehicles.pdf}.

\bibitem{ichnowski2022fogros}
J.~Ichnowski, K.~Chen, K.~Dharmarajan, S.~Adebola, M.~Danielczuk,
  V.~Mayoral-Vilches, H.~Zhan, D.~Xu, R.~Ghassemi, J.~Kubiatowicz
  \emph{et~al.}, ``Fogros 2: An adaptive and extensible platform for cloud and
  fog robotics using ros 2,'' \emph{arXiv preprint arXiv:2205.09778}, 2022.

\bibitem{ros2-timer}
``{ROS}2 {W}all{T}imer,''
  \url{http://docs.ros.org/en/indigo/api/roscpp/html/classros_1_1WallTimer.html}.

\bibitem{kumar2012cloud}
S.~Kumar, S.~Gollakota, and D.~Katabi, ``A cloud-assisted design for autonomous
  driving,'' in \emph{Proceedings of the first edition of the MCC workshop on
  Mobile cloud computing}, 2012, pp. 41--46.

\bibitem{qiu2018avr}
H.~Qiu, F.~Ahmad, F.~Bai, M.~Gruteser, and R.~Govindan, ``Avr: Augmented
  vehicular reality,'' in \emph{Proceedings of the 16th Annual International
  Conference on Mobile Systems, Applications, and Services}, 2018, pp. 81--95.

\bibitem{chen2019f}
Q.~Chen, X.~Ma, S.~Tang, J.~Guo, Q.~Yang, and S.~Fu, ``F-cooper: Feature based
  cooperative perception for autonomous vehicle edge computing system using 3d
  point clouds,'' in \emph{Proceedings of the 4th ACM/IEEE Symposium on Edge
  Computing}, 2019, pp. 88--100.

\bibitem{chinchali2019neural}
S.~P. Chinchali, E.~Cidon, E.~Pergament, T.~Chu, and S.~Katti, ``Neural
  networks meet physical networks: Distributed inference between edge devices
  and the cloud,'' in \emph{Proceedings of the 17th ACM Workshop on Hot Topics
  in Networks}, 2018, pp. 50--56.

\bibitem{zhang2021emp}
X.~Zhang, A.~Zhang, J.~Sun, X.~Zhu, Y.~E. Guo, F.~Qian, and Z.~M. Mao, ``Emp:
  Edge-assisted multi-vehicle perception,'' in \emph{Proceedings of the 27th
  Annual International Conference on Mobile Computing and Networking}, 2021,
  pp. 545--558.

\bibitem{kang2017neurosurgeon}
Y.~Kang, J.~Hauswald, C.~Gao, A.~Rovinski, T.~Mudge, J.~Mars, and L.~Tang,
  ``Neurosurgeon: Collaborative intelligence between the cloud and mobile
  edge,'' \emph{ACM SIGARCH Computer Architecture News}, vol.~45, no.~1, pp.
  615--629, 2017.

\bibitem{cui2020offloading}
M.~Cui, S.~Zhong, B.~Li, X.~Chen, and K.~Huang, ``Offloading autonomous driving
  services via edge computing,'' \emph{IEEE Internet of Things Journal},
  vol.~7, no.~10, pp. 10\,535--10\,547, 2020.

\bibitem{sun2018learning}
Y.~Sun, X.~Guo, S.~Zhou, Z.~Jiang, X.~Liu, and Z.~Niu, ``Learning-based task
  offloading for vehicular cloud computing systems,'' in \emph{2018 IEEE
  International Conference on Communications (ICC)}.\hskip 1em plus 0.5em minus
  0.4em\relax IEEE, 2018.

\bibitem{sun2018cooperative}
F.~Sun, F.~Hou, N.~Cheng, M.~Wang, H.~Zhou, L.~Gui, and X.~Shen, ``Cooperative
  task scheduling for computation offloading in vehicular cloud,'' \emph{IEEE
  Transactions on Vehicular Technology}, vol.~67, no.~11, pp. 11\,049--11\,061,
  2018.

\bibitem{stoica2021cloud}
I.~Stoica and S.~Shenker, ``From cloud computing to sky computing,'' in
  \emph{Proceedings of the Workshop on Hot Topics in Operating Systems}, 2021,
  pp. 26--32.

\bibitem{chameleon}
\BIBentryALTinterwordspacing
J.~Jiang, G.~Ananthanarayanan, P.~Bodik, S.~Sen, and I.~Stoica, ``{Chameleon:
  Scalable Adaptation of Video Analytics},'' in \emph{Proceedings of the ACM
  Special Interest Group on Data Communication Conference (SIGCOMM)}, 2018, pp.
  253--266. [Online]. Available:
  \url{http://doi.acm.org/10.1145/3230543.3230574}
\BIBentrySTDinterwordspacing

\bibitem{cruise-welcome-riders}
``Welcome, riders,''
  \url{https://getcruise.com/news/blog/2022/welcome-riders/}, Aug. 2021.

\bibitem{waymo-sf-riders}
``Welcoming our first riders in san francisco,''
  \url{https://blog.waymo.com/2021/08/welcoming-our-first-riders-in-san.html},
  Feb. 2022.

\bibitem{grpc}
``grpc,'' \url{https://grpc.io/}.

\bibitem{youtube-bitrate}
``Youtube recommended upload encoding settings,''
  \url{https://support.google.com/youtube/answer/1722171}.

\bibitem{carla}
A.~Dosovitskiy, G.~Ros, F.~Codevilla, A.~Lopez, and V.~Koltun, ``{CARLA: An
  Open Urban Driving Simulator},'' in \emph{Proceedings of the
  1\textsuperscript{st} Conference on Robot Learning (CoRL)}, 2017, pp. 1--16.

\bibitem{carla-nhtsa-scenarios}
``{NHTSA-inspired Pre-crash Scenarios},''
  \url{https://carlachallenge.org/challenge/nhtsa/}.

\bibitem{nhtsa-precrash-scenarios}
``{Pre-Crash Scenario Typology for Crash Avoidance Research},''
  \url{https://www.nhtsa.gov/sites/nhtsa.gov/files/pre-crash_scenario_typology-final_pdf_version_5-2-07.pdf}.

\end{thebibliography}
}

\end{document}